# Efficient Edge LLMs Deployment via Hessian-Aware Quantization and CPU–GPU Collaborative

Tuo Zhang, Ning Li, Xin Yuan, Wenchao Xu, Quan Chen, Song Guo, *Fellow, IEEE*, Haijun Zhang, *Fellow, IEEE*

*Abstract*—With the breakthrough progress of large language models (LLMs) in natural language processing and multimodal tasks, efficiently deploying them on resource-constrained edge devices has become a critical challenge. The Mixture-of-Experts (MoE) architecture enhances model capacity through sparse activation, but faces two major difficulties in practical deployment: (1) The presence of numerous outliers in activation distributions leads to severe degradation in quantization accuracy for both activations and weights, significantly impairing inference performance; (2) Under limited memory, efficient offloading and collaborative inference of expert modules struggle to balance latency and throughput. To address these issues, this paper proposes an efficient MoE edge deployment scheme based on Hessian-Aware Quantization (HAQ) and CPU-GPU collaborative inference. First, by introducing smoothed Hessian matrix quantization, we achieve joint 8-bit quantization of activations and weights, which significantly alleviates the accuracy loss caused by outliers while ensuring efficient implementation on mainstream hardware. Second, we design an expert-level collaborative offloading and inference mechanism, which, combined with expert activation path statistics, enables efficient deployment and scheduling of expert modules between CPU and GPU, greatly reducing memory footprint and inference latency. Extensive experiments validate the effectiveness of our method on mainstream large models such as the OPT series and Mixtral-8×7B: on datasets like Wikitext2 and C4, the inference accuracy of the low-bit quantized model approaches that of the full-precision model, while GPU memory usage is reduced by about 60%, and inference latency is significantly improved. This work provides a practical technical solution and engineering reference for the efficient and stable deployment of large-scale MoE models in real-world edge environments.

*Index Terms*—edge deployment, CPU-GPU collaboration, Hessian-aware quantization, Mixture of Experts, quantization.

## I. INTRODUCTION

WITH the rapid advancements of Large Language Models (LLMs) in natural language processing (NLP) and multimodal perception (e.g., [1], [2], [26]), efficiently deploying these models on edge devices has emerged as a significant area of research at the intersection of artificial intelligence and edge computing. Edge deployment of LLMs dramatically reduces data transmission latency and enables real-time, on-device inference, making it particularly suitable for applications that demand low latency and high data security—such as smart terminals, industrial automation, intelligent manufacturing, automotive systems, healthcare, and security surveillance.

By shifting computation closer to the data source, edge-based solutions not only improve responsiveness but also address growing privacy and regulatory concerns that accompany large-scale AI deployments.

Compared to conventional cloud-based inference approaches, performing inference at the edge enables data collection, preprocessing, and model execution to occur locally. This significantly reduces network communication overhead, leading to faster system response times and more reliable service continuity. Moreover, processing data at its source minimizes the risk of sensitive information being exposed during transmission or storage, thereby strengthening privacy protection and data security, and ensuring compliance with evolving regulations such as GDPR. Edge devices are also equipped with autonomous decision-making capabilities and robust fault tolerance mechanisms [31], allowing them to independently perform critical tasks even under unreliable network conditions or during cloud service interruptions. This autonomy is vital for the availability and resilience of intelligent systems.

While foundational methods such as model quantization (e.g., [6], [7], [9], [10], [11]), pruning (e.g., [29]), knowledge distillation (e.g., [28]), and low-rank decomposition (e.g., [27]) have made edge deployment feasible, and collaborative inference and model offloading techniques have been introduced to leverage heterogeneous computing resources (e.g., CPUs and GPUs) for improved throughput and responsiveness, these solutions still face challenges. Specifically, they may not fully satisfy the stringent requirements for efficiency, low latency, and accuracy demanded by large-scale LLM inference at the edge.

The Mixture-of-Experts (MoE) architecture has emerged as a leading solution for scaling Large Language Models (LLMs), owing to its sparse activation and strong scalability. However, deploying MoE models on edge devices introduces several critical challenges that must be overcome to achieve efficient and reliable inference.

First, quantizing both activations and weights in MoE models often leads to significant accuracy loss, primarily due to the prevalence of outliers in activation distributions. This issue is particularly pronounced in low-bit quantization (such as INT8), where outlier values expand the quantization range, causing most activation values to be poorly represented and thus amplifying quantization errors. As a result, the inference accuracy can drop sharply. Three main issues arise in quantizing LLM activations: (1) activation distributions are highly skewed, with outliers dominating and reducing effective bit utilization; (2) these outliers are typically concentrated in a few stable channels, but mainstream GEMM kernels generally lack support for per-channel quantization, resulting in reliance on per-tensor or per-token quantization schemes, which are less effective at mitigating outlier impact; (3) the persistence of outliers suggests that adaptive scaling could be beneficial, yet existing solutions provide limited adaptability. For example, SmoothQuant [8] introduces a smoothing factor for joint quantization but typically relies on empirically set, static parameters, leading to inconsistent performance across diverse scenarios.

Second, the dynamic and complex activation patterns observed in MoE models pose significant obstacles to effective expert offloading and collaborative inference. Most current offloading and inference strategies employ coarse-grained approaches, which fail to fully leverage heterogeneous computing resources such as CPUs and GPUs. This limitation results in increased system latency, reduced throughput, and suboptimal cache hit rates. Moreover, frequent migration of expert parameters between CPUs and GPUs contributes to higher data transfer overhead and greater variability in inference latency—issues that are particularly detrimental in real-time edge applications.

Third, efficiently caching frequently activated experts within the limited memory resources of a GPU is essential for improving cache hit rates and minimizing cross-device transfer latency. However, existing strategies for expert cache management, replacement, and activation prediction remain inadequate, often yielding unstable performance under high concurrency or complex workload conditions.

Finally, there is a lack of end-to-end solutions that seamlessly integrate efficient quantization with dynamic expert offloading and collaborative inference. In practice, quantization-induced accuracy loss and the computational challenges of dynamic expert scheduling are closely intertwined. Without coordinated optimization, the advantages of the MoE architecture for edge deployment cannot be fully realized and may even exacerbate system performance bottlenecks.

To address these challenges, we propose an efficient Mixture-of-Experts (MoE) model inference framework tailored for edge devices, introducing innovations at both the model compression and expert scheduling levels.

First, we present a Hessian-Aware Quantization (HAQ) method for joint quantization:Our approach overcomes the limitations of conventional quantization methods in handling activation outliers by introducing a two-level optimization strategy that significantly enhances quantization accuracy and model robustness.At the activation level, we employ an optimized search algorithm to dynamically determine the activation smoothing factor, replacing the empirical parameter selection used in SmoothQuant. This adaptive process improves robustness across diverse input scenarios.At the weight level, drawing inspiration from GPTQ [6], we incorporate Hessian-based perturbation estimation and residual compensation to achieve weight quantization while preserving high-precision outputs.Furthermore, we extend this approach by developing a precision-heterogeneous quantization deployment strategy to better exploit heterogeneous computing resources on edge devices.

Second, we propose an expert-level CPU–GPU collaborative inference scheme:This includes the systematic optimization of expert offloading, scheduling, and caching mechanisms based on the observed distribution patterns of expert activation pathways. (1) We introduce a fine-grained expert offloading mechanism, whereby frequently activated experts and shared modules remain resident on the GPU, while less frequently used experts are migrated to the CPU. (2) A runtime cost predictor is employed to dynamically determine whether activations are computed on the CPU or whether expert parameters should be offloaded to the GPU, thereby optimizing overall system latency. (3) We design a GPU-side expert caching mechanism that utilizes a Least Recently Used (LRU) policy to manage cache states, improving expert cache hit rates and reducing cross-device transfer overhead.Additionally, a staged expert deployment strategy is introduced, which holistically considers both the frequency of expert activation paths and the global popularity of individual experts. This strategy prioritizes high-frequency path experts, followed by those with generally high activation frequency, thereby achieving an optimal balance of cache efficiency across layers and significantly enhancing inference stability and system throughput.

Experimental results show that our approach achieves near-FP16 accuracy under 8-bit quantization on models such as OPT series [30] and Mixtral-8×7B [5] , on the Wikitext2 and C4 datasets, while reducing memory usage by about 60%. At the same time, the heterogeneous inference system demonstrates higher expert hit rates, lower inference latency fluctuations, and stronger robustness, providing a practical technical path and engineering reference for the efficient deployment of large-scale MoE models on edge devices.

## II. RELATED WORKS

In recent years, Large Language Models (LLMs) have achieved remarkable success across a wide range of natural language processing tasks, driving a growing demand for their deployment on edge devices. However, the limited computational and memory resources available on such devices pose significant challenges for efficient LLM inference, making this an active area of research. In this section, we provide a review of related work most pertinent to our study.

### A. Model Quantization Methods

Mainstream neural network quantization methods can be broadly classified into Post-Training Quantization (PTQ) and Quantization-Aware Training (QAT). PTQ approaches have garnered significant attention in recent years, as they circumvent the need for model retraining, thereby substantially reducing both computational and time costs. This advantage is particularly pronounced for large-scale pre-trained models with billions of parameters, such as Large Language Models (LLMs). In contrast, QAT methods enhance model robustness to quantization by incorporating quantization noise during training, but they typically require considerable computational resources and extensive training data. As a result, QAT is often impractical for large models due to its high deployment cost.

Among PTQ methods, GPTQ [6] stands out by utilizing the Hessian matrix of the weights—approximated via activation values—to achieve fine-grained control over quantization error. GPTQ employs a block-wise, layer-wise quantization strategy, partitioning large models into smaller blocks and sequentially quantizing and calibrating each block. This approach effectively mitigates the accumulation of global quantization errors. Furthermore, GPTQ incorporates engineering optimizations such as groupwise batch updates and streamlined matrix operations, greatly enhancing quantization efficiency. Consequently, it becomes feasible to compress billion-parameter models to low-bit formats (e.g., 4-bit or 8-bit) with high precision on a single GPU. Experimental evidence demonstrates that GPTQ can substantially reduce model storage and computational demands while preserving inference accuracy, offering a practical solution for deploying LLMs in resource-constrained environments such as edge devices.

In addition, AWQ [7] advances weight quantization by addressing channel sensitivity through an innovative channel importance analysis. AWQ first identifies critical weight channels that exert significant influence on model outputs, and applies protective rescaling to these channels during quantization. This targeted approach effectively mitigates the accumulation of quantization errors along critical computational paths. Compared to GPTQ, AWQ eliminates the need for backpropagation or additional fine-tuning, further reducing the computational and engineering overhead associated with the quantization process. Empirical evaluations demonstrate that AWQ delivers superior quantization accuracy and inference performance on a range of mainstream large models, such as LLaMA and OPT. However, it is important to note that mainstream PTQ methods like GPTQ and AWQ primarily focus on weight quantization. Their effectiveness in scenarios requiring joint quantization of both activations and weights remains limited, making it challenging to support unified low-bit operators on certain edge and heterogeneous hardware platforms.

To address these limitations, SmoothQuant [8] introduces a novel approach that mathematically shifts the quantization difficulty from activations to weights. Specifically, SmoothQuant normalizes and smooths activations prior to quantization and applies weight reconstruction techniques, enabling joint quantization of both weights and activations to 8 bits (W8A8) while maintaining high inference performance. This method demonstrates strong generality and robustness across diverse model architectures and downstream tasks, achieving performance close to full-precision models on most mainstream NLP benchmarks. Moreover, SmoothQuant significantly enhances deployment efficiency on edge devices. Nevertheless, there remains room for improvement in certain scenarios, and its adaptability to ultra-low-bit quantization (e.g., 4-bit) requires further investigation.

Furthermore, OmniQuant [10] extends joint quantization by introducing learnable equivalent transformation and clipping mechanisms. By incorporating learnable parameter adjustment strategies during quantization, OmniQuant dynamically optimizes the distributions of weights and activations, enabling more flexible trade-offs between quantization precision and inference accuracy. This approach not only broadens the applicability of quantization techniques but also achieves notable improvements in model performance. Unlike traditional PTQ methods, OmniQuant introduces limited parameter fine-tuning during the quantization process, effectively alleviating the accuracy degradation typically associated with joint quantization of activations and weights. Experimental results indicate that OmniQuant achieves strong quantization performance across a variety of mainstream models and tasks. However, its applicability to specialized architectures such as Mixture-of-Experts (MoE) remains limited, and the additional training required increases computational demands, necessitating further optimization for large-scale, real-world deployments.

In summary, existing PTQ and QAT approaches provide a robust theoretical and methodological foundation for the efficient deployment of high-performance large language models on resource-constrained devices. PTQ methods, owing to their efficiency and low computational overhead, have become the mainstream choice for large model quantization, while QAT offers enhanced solutions for scenarios demanding the highest levels of accuracy. Nonetheless, key challenges remain in the joint quantization of activations and weights, adaptation to specialized model architectures (such as Mixture-of-Experts and Transformer variants), and achieving extreme performance optimization. Future research should prioritize improving the generality of quantization techniques, developing end-to-end automated quantization pipelines, and optimizing adaptation to heterogeneous hardware platforms. These advances are essential for further enabling the widespread deployment and practical application of large-scale language models in resource-constrained environments such as edge computing.

### B. Model Offloading and Heterogeneous Collaborative Inference

To address memory bottlenecks and enable deployment in resource-constrained environments, model offloading techniques have emerged as a critical strategy for extending the inference capabilities of large models. The core principle involves offloading a portion of model weights or computational tasks from GPU to CPU memory—or even to disk storage—and dynamically reloading data to the GPU as needed. This approach effectively overcomes the capacity limitations of a single GPU, thereby supporting inference for models of greater scale.

Conventional model offloading strategies are typically GPU-centric, utilizing CPU memory as auxiliary storage. In practice, parameters that exceed GPU memory capacity are stored in CPU memory and transferred to the GPU on demand during inference, as illustrated in **Fig. 1**. Provided that sufficient CPU memory and disk space are available, this method can, in theory, support inference for LLMs of virtually any size [12], [13], [20], [21], [24]. However, this strategy introduces significant performance bottlenecks: frequent data transfers between the GPU and CPU during inference—particularly in token-by-token autoregressive generation scenarios (e.g., batch size of 1)—make data transfer latency the dominant limiting factor. In such cases, more than 99.5% of processing time may be consumed by weight transfers, severely degrading overall inference latency and negatively impacting user experience.

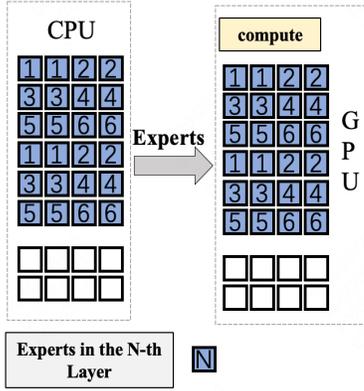

**Fig. 1.** GPU-centric Hybrid model offloading

To overcome these limitations, hybrid offloading strategies have been developed, wherein model parameters are partitioned between the GPU and CPU based on functional roles or structural components. In these approaches, certain layers or expert modules reside in CPU memory, while the remainder remain on the GPU, as illustrated in **Fig. 2**. During inference, the CPU not only serves as a storage medium for parameters but can also directly participate in computation, passing intermediate activations or results to the GPU for subsequent processing. This design effectively reduces the volume of data transferred per inference step, alleviates PCIe bandwidth constraints, and significantly lowers inference latency.

Building upon this foundation, recent research has explored heterogeneous collaborative inference architectures that further enhance the efficiency and scalability of large model inference through finer-grained task partitioning and dynamic scheduling. For example, Fiddler [14] implements an inference strategy based on activation transfer, fully leveraging CPU computational capabilities to share the expert module workload, thereby enabling inference of MoE models with over 90 billion parameters on a 40GB GPU. EdgeMoE [12] introduces expert-level bit-width adaptation and preloading mechanisms, coupled with I/O access prediction, to optimize model weight loading and improve expert cache hit rates and overall inference efficiency. PowerInfer [15], designed for dense models, proposes a CPU–GPU collaborative inference framework based on neuron "hotness" and operator adaptation, achieving substantial performance improvements on edge devices—an approach that is also extensible to MoE models.

Overall, model offloading technologies offer a promising solution for deploying large models on memory-constrained devices, while CPU–GPU collaborative strategies demonstrate significant potential for reducing inference latency, increasing throughput, and improving resource utilization. However, current model offloading methods are not yet effectively integrated with quantization techniques. In summary, model offloading expands deployment options for large models in memory-limited scenarios, and CPU–GPU collaboration further advances latency control and throughput optimization.

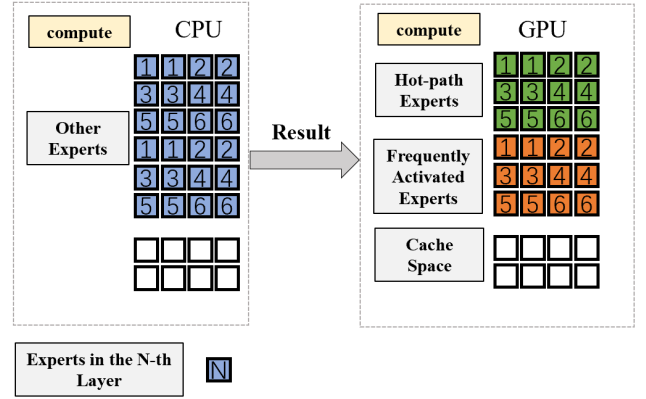

**Fig. 2.** We propose a hybrid offloading scheme for CPU-GPU collaborative inference.
(The figure assumes the MoE model has 6 layers, each with 8 experts.)

## III. EFFICIENT EDGE DEPLOYMENT OF MIXTURE-OF-EXPERTS LLMS

This section focuses on two critical challenges associated with deploying Mixture-of-Experts (MoE) models on edge devices. First, there is a lack of efficient quantization techniques capable of jointly quantizing both activations and weights in the context of MoE deployment. Second, while model offloading is widely adopted to address memory constraints on edge devices, there remain few inference solutions that are both compatible with quantization methods and capable of delivering high efficiency.

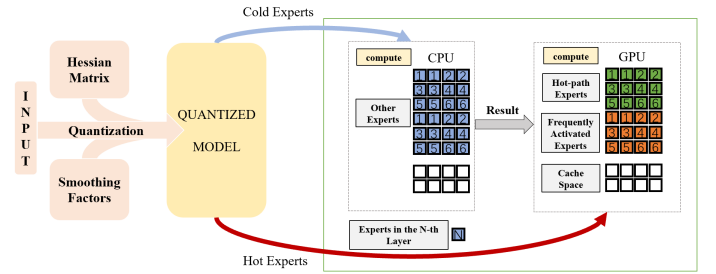

**Fig. 3.** The overall architecture.

### A. Overview

To address these challenges, we propose two complementary solutions: (1) We introduce a smoothed Hessian matrix-based quantization mechanism with importance ranking, which transfers the quantization difficulty and enables more accurate joint quantization of activations and weights; (2) We develop an expert-distributed, CPU–GPU collaborative inference strategy specifically designed for MoE models.

The activation–weight quantization module substantially reduces the parameter storage requirements of MoE models, while the CPU–GPU collaborative inference module enables fine-grained partitioning and deployment of experts across heterogeneous resources. Together, these modules overcome the memory bottleneck inherent to MoE architectures and facilitate fast, memory-efficient inference without sacrificing model accuracy. The overall framework is illustrated in **Fig. 3.**

### B. Activation-Weight Hessian-Aware Quantization (HAQ)

#### 1) Optimized Smoothing Factor

As mentioned earlier, the challenge of activation quantization is much greater in large language models (LLMs) than in small- and medium-scale neural networks. The core difficulty lies in the prevalence of outliers within the activation distribution. Extreme outliers in the activations can greatly expand the quantization range, causing the vast majority of activation values to cluster in the middle of the quantization interval and be coarsely represented. Ultimately, this substantially amplifies quantization error and leads to a sharp drop in inference accuracy after low-bit quantization (such as INT8 or lower). Outliers in activation $X$ cause the quantization interval to be excessively stretched, leaving very few effective bits for most regular activation values—a phenomenon particularly prominent in LLMs and a primary barrier to efficient activation-weight quantization.

Traditional methods introduce a smoothing factor to jointly scale activations and weights, making the activation distribution more concentrated and thereby reducing the impact of outliers on the quantization range. The smoothing factor $s$ is chosen to minimize the error between the quantized and original outputs:

$$s = \min_s \| Q(W \cdot s)Q(s^{-1} \cdot X) - WX \| \qquad (1)$$

where $Q(\cdot)$ denotes the quantization operation, $W$ is the weight, and $X$ is the activation. This approach helps suppress quantization errors caused by outliers and improves the performance of low-bit models. However, the smoothing factor is usually set empirically and lacks adaptability, which may lead to unstable quantization results across different models or data distributions.

To overcome this limitation, this paper draws on the ideas of AWQ in weight quantization and extends them to handle activation outliers. The core of AWQ lies in channel-level importance analysis and adaptive scaling. Specifically, AWQ analyzes the importance of weight channels and uses per-channel scaling factors to protect critical channels, significantly reducing quantization error. Experimental findings indicate that, compared to traditional methods of selecting significant channels based on weight norms, protecting weight channels according to the magnitude of the corresponding activation channels can more effectively improve quantization accuracy. This observation reveals a close relationship between the magnitude of activation channels and the importance of their corresponding weight channels: larger activation channels have a greater impact on overall accuracy during quantization and thus deserve special attention.

Building on this, this paper innovatively extends AWQ's optimization objective from weight quantization to joint weight-activation quantization. Specifically, the objective is adjusted from (2) to (1):

$$s = \min_s \| Q(W \cdot s)(s^{-1} \cdot X) - WX \| \qquad (2)$$

Joint optimization of activations and smoothing factors can more effectively reduce quantization error and further improve inference accuracy under low-bit quantization. To achieve efficient and adaptive smoothing, we propose a grid-search-based scaling factor optimization method. Specifically, we perform a grid search for the scaling factor $e$ within the interval $[0,1]$ to determine the optimal scaling parameter. The smoothing factor $s$ is then determined as follows, depending only on the magnitude of the activation $X$:

$$s = s_X^e \qquad (3)$$

This approach eliminates dependence on empirical settings and instead uses a data-driven method to adaptively determine scaling and smoothing parameters, thereby more effectively suppressing the negative impact of activation outliers on quantization accuracy. By jointly optimizing the quantization of weights and activations, the error between the quantized and original outputs is effectively reduced, improving the inference accuracy and stability of LLMs under low-bit quantization.

Moreover, our method takes hardware implementation feasibility into account. Given the limited support for per-channel scaling in mainstream high-throughput GEMM kernels, the proposed scaling strategy enables effective smoothing of activation outliers without significantly increasing computational or memory overhead. By applying scaling on external dimensions (such as the token dimension and output channel $C_o$ ), our approach ensures both quantization accuracy and inference efficiency, making it suitable for deploying large-scale models on real-world edge devices.

In summary, the proposed method offers the following advantages:

a) It can automatically adapt to different model architectures and data distributions, significantly improving the robustness and generalization of quantization;

b) By jointly optimizing the quantization of weights and activations, it effectively reduces quantization error and significantly improves inference accuracy under low-bit quantization;

c) Through efficient optimization techniques such as grid search, it balances quantization accuracy and efficiency, making it suitable for practical deployment of large-scale models.

In conclusion, this adaptive smoothing and joint optimization strategy addresses the challenges posed by activation outliers in LLM quantization, providing a new theoretical foundation and engineering pathway for improving quantized inference performance of large models in resource-constrained environments, and laying a solid foundation for the future efficient low-bit deployment of large-scale neural network models.

*2) Hessian Matrix-Based Weight Quantization*

After activation smoothing, the distribution of model activations becomes significantly more concentrated, effectively suppressing the stretching effect of extreme outliers on the overall distribution. This transformation establishes a solid foundation for subsequent efficient activation–weight quantization. As a result, the majority of activation values are confined within a narrower interval, the influence of outlier channels on the quantization range is greatly diminished, and the model's overall quantization error is substantially reduced. With this optimized activation distribution, the performance upper bound for the subsequent weight quantization stage can be further elevated.

For weight quantization, one of the most representative and effective approaches is GPTQ. The core innovation of GPTQ lies in the incorporation of second-order information—specifically, leveraging the sensitivity of model outputs to weight perturbations as captured by the Hessian matrix to guide quantization optimization. Unlike traditional uniform quantization or methods that merely minimize weight distortion, Hessian-based quantization explicitly formulates the quantization objective as minimizing the mean squared error (MSE) between the model output after quantization and the original full-precision output. By progressively optimizing each row of the weight matrix, this method enables rational allocation of quantization error, thereby maximally preserving the inference capability of the network.

Specifically, the optimization objective for weight quantization can be formalized as follows: for a given input activation X, the goal is to find a quantized weight matrix $\widehat{W}$ such that the ℓ2 norm distance between the quantized model output $|\widehat{W}X|$ and the full-precision output $|WX|$ is minimized:

$$\underset{\widehat{W}}{\operatorname{argmin}} \left\| WX - \widehat{W}X \right\|_2^2 \qquad (4)$$

In practice, a row-wise quantization strategy is adopted. When quantizing each row of weights, since the objective is a quadratic optimization problem, the corresponding Hessian matrix $H_F$ can be efficiently computed using the input activations $X_F$ as : $H_F = 2X_F X_F^T$.

where F denotes the set of weights not yet quantized. Suppose the current weight to be quantized is wi, the Hessian-based quantization proceeds via the following steps:

1. Selection of the Optimal Quantized Value

For each weight wi to be quantized, GPTQ selects the quantized value $Q(w_i)$ that minimizes the objective function as follows:

$$w_i = argmin_{w_i} \frac{(Q(w_i) - w_i)^2}{[H_F^{-1}]_{ii}} \qquad (5)$$

where $[H_F^{-1}]_{ii}$ is the i-th diagonal element of the inverse Hessian matrix, reflecting the impact of perturbing the current weight on the overall output error. This design adaptively allocates quantization error according to each weight's sensitivity to the final output, prioritizing the protection of weights with greater influence.

2. Weight Compensation and Error Propagation

Quantizing wi introduces disturbance to the model output. To prevent error accumulation in subsequent quantization steps, a weight compensation mechanism is introduced. By computing a compensation vector $\delta_F$, the remaining unquantized full-precision weights are adjusted to balance the error introduced by already quantized weights. The compensation update formula is:

$$\delta_F = -\frac{w_i - Q(w_i)}{[H_F^{-1}]_{ii}} \cdot (H_F^{-1})_{:,i} \qquad (6)$$

Where $(H_F^{-1})_{:,i}$ is the i-th column of the inverse Hessian matrix. This compensation mechanism reasonably distributes the quantization error among the unquantized weights, suppressing the cascading effect of errors.

3. Iterative Quantization and Recursive Updates

Through the above optimal quantization and compensation steps, each row of weights is quantized and error-adjusted in turn, until all weights are quantized. Throughout the process, the inverse Hessian matrix can be efficiently updated using methods such as Cholesky reformulation, ensuring computational efficiency and controllable memory usage, which is suitable for large-scale model compression in practice.

Overall, the weight quantization process can be summarized as follows:

a)  Evaluate the perturbation caused by quantizing each row of weights based on the Hessian matrix, fully considering the correlation and global impact among weights;

b)  Minimize the impact of quantization error on model output and protect inference accuracy through optimal quantized value selection and compensation mechanisms;

c)  Employ efficient recursive update strategies to ensure scalability and efficiency on large models.

Compared to traditional quantization methods, Hessian-based quantization offers substantial improvements in the performance of large models under low-bit quantization, making it particularly well suited for the efficient deployment of large-scale Transformer models in resource-constrained environments such as edge devices. When combined with the aforementioned activation smoothing strategy, joint activation–weight quantization enables further compression of model size, reduction in computational resource consumption, and optimization of both inference efficiency and energy usage—while maintaining high model accuracy. The synergistic integration of these techniques provides a robust theoretical foundation and practical technical support for the deployment of large language models in real-world engineering applications.

*3) Device-Aware Heterogeneous Precision Adaptation*

During the edge deployment of large-scale Mixture-of-Experts (MoE) models, a central challenge is how to fully exploit the complementary hardware capabilities of heterogeneous computing platforms—such as CPUs and GPUs—to maximize inference performance and resource utilization. The massive parameter scale, numerous expert modules, and complex, dynamic activation patterns of MoE models often exceed the capabilities of any single type of computing unit to deliver both efficiency and flexibility. As a result, precision-heterogeneous parameter management and inference strategies that leverage both CPUs and GPUs provide a practical pathway for efficient deployment of large models in resource-constrained environments.

CPUs and GPUs exhibit fundamentally complementary characteristics in terms of architecture and resource configuration. CPUs typically offer large-capacity main memory (RAM) capable of storing full-precision model parameters (e.g., FP16 or FP32) and supporting high-bandwidth data access, making them well suited for memory-intensive operations such as model loading and cache management. In contrast, GPUs excel at parallel computation and high-throughput matrix multiplication (GEMM), but are limited by relatively constrained on-chip memory (VRAM) and bandwidth, making it impractical to store all full-precision model parameters directly on the GPU. Consequently, precision-heterogeneous parameter management—tailored to the strengths of each hardware platform—is essential for improving inference efficiency and overall system throughput.

Specifically, on the CPU side, the ample memory resources are utilized to store model weights in a compressed low-bit format (such as INT8) during model loading and parameter management. Prior to inference, the CPU performs dequantization, converting these quantized weights to full-precision formats (e.g., FP16 or FP32) and caching them in memory to facilitate efficient expert module computation. By performing dequantization once during model loading, this approach avoids repeated low-bit to high-precision decoding during inference, significantly reducing runtime overhead on the CPU.

On the GPU side, inference proceeds by directly loading expert weights in INT8 format and employing optimized low-precision GEMM kernels for high-throughput matrix operations. This low-precision inference paradigm not only reduces VRAM consumption and alleviates bandwidth constraints, but also fully leverages the computational power of the GPU, resulting in higher throughput and concurrency. With a carefully designed quantization strategy, INT8 inference can outperform traditional full-precision inference in both speed and energy efficiency under controlled accuracy loss, making it especially advantageous for edge devices with stringent real-time and energy efficiency requirements.

Moreover, the precision-heterogeneous management scheme supports flexible expert module scheduling. Given the dynamic activation patterns of expert modules in MoE inference, the system can rapidly select and transfer the required experts' weights—stored in low-bit format—to the GPU for efficient computation, based on real-time input and scheduling strategies. This mechanism effectively reduces data transfer latency and mitigates bandwidth bottlenecks.

Throughout the inference process, the division of labor between CPU and GPU is clear: the CPU manages parameters, performs pre-dequantization, and assists in inference computation, while the GPU is dedicated to compute-intensive matrix operations. This collaborative approach prevents CPU bottlenecks and relieves VRAM pressure on the GPU, achieving optimal resource allocation and utilization. For large-scale MoE deployment on edge devices, this strategy significantly boosts inference throughput and overall system performance while maintaining inference quality.

In summary, precision-heterogeneous model management and inference strategies—grounded in the hardware characteristics of CPUs and GPUs—enable the seamless integration of quantization techniques with edge-distributed deployment. This approach not only facilitates efficient and cost-effective deployment of large models on edge devices, but also establishes a theoretical and engineering paradigm for intelligent inference systems involving multi-device collaboration and dynamic expert scheduling. Such solutions hold significant promise for enhancing the performance of edge AI systems, reducing energy consumption, and lowering deployment costs.

The specific steps of HAQ can be found in **Algorithm 1**.

---

**Algorithm 1** Activation-Weight Hessian-Aware Quantization (HAQ)

| | | |
|---|---|---|
| Input:Model | $\longrightarrow$ | Pretrained LLM (with MoE) |
| X_calib | $\longrightarrow$ | Calibration activation data |
| BitW, BitA | $\longrightarrow$ | Quantization bits for weights/activations |
| device_type | $\longrightarrow$ | {'CPU', 'GPU'} |
| Output:Quantized_Model | $\longrightarrow$ | Model ready for heterogeneous deployment |

Procedure:

1. **Adaptive Activation Smoothing**
   for each layer in Model:
       X_layer ← collect_activations(X_calib, layer)
       best_loss ← ∞
       for e in grid_search_range(0, 1, step):
           s ← (abs(X_layer))^e
     → Channel-wise smoothing
           X_smooth ← X_layer / s
           W_smooth ← layer.weight * s
           loss ←
   evaluate_quantization_loss(W_smooth, X_smooth)
           if loss < best_loss:
               best_e ← e
               best_s ← s
               best_loss ← loss
       layer.smoothing_factor ← best_s
       layer.weight ← layer.weight * best_s
       layer.activation ← layer.activation / best_s

2. **Hessian-Aware Weight Quantization**
   for each layer in Model:
       X_calib ← collect_activations(X_calib, layer)
       W ← layer.weight
       H ← 2 × X_calib × X_calib^T
     → Hessian matrix
       H_inv ← inverse(H)
       for i in range(num_rows(W)):
           q_w ← quantize_row(W[i], BitW)
           q_w_opt ← argmin_over_q ( (q_w - W[i])^2 / H_inv[i, i] )

delta ← -(W[i] - q_w_opt) / H_inv[i, i] ×
H_inv[:, i]
⟶ Compensation
    W[not_quantized] ← W[not_quantized] +
delta[not_quantized]
    W[i] ← q_w_opt
    layer.weight ← W

3. **Device-Aware Precision Adaptation**
    **for** each expert **in** Model.experts:
      **if** device_type == 'CPU':
        expert.weight_fp ←
dequantize(expert.weight_int, target_dtype='fp16')
        cache_in_cpu(expert.weight_fp)
      **elif** device_type == 'GPU':
        expert.weight_int ← quantize(expert.weight,
BitW)
        cache_in_gpu(expert.weight_int)

4. **Return** Quantized Model

---

*C. CPU–GPU Collaborative Inference*

*1) GPU-CPU Hybrid Model Offloading*

Current model compression techniques remain inadequate for deploying large language models (LLMs) on resource-limited consumer GPUs. As a result, offloading computation and storage to the CPU presents a more practical and scalable solution. In this section, we introduce a collaborative framework comprising three key modules to enable efficient GPU–CPU hybrid model offloading, as illustrated in **Fig. 2.**

*a.Cost-Benefit Analysis Foundation for Offloading Strategy:*

Although the peak computational power of the CPU is generally much lower than that of the GPU, the key decision in expert offloading is to compare the total latency of two options:

- **Option A (CPU computation):** Compute the selected expert's output directly on the CPU, then transfer the resulting intermediate activation back to the GPU for subsequent processing. The total latency $T_{CPU}$ is primarily determined by the CPU computation time.

- **Option B (Transfer to GPU for computation):** Transfer the selected expert's parameters from CPU memory to GPU memory, then perform the computation on the GPU. The total latency $T_{sum}$ is the sum of the expert parameter transfer time $T_e$ and the GPU computation time $T_{GPU}$

In practice, since expert parameters are typically large—especially for LLMs—the transfer time $T_e$ can be substantial, often making $T_{sum}$ significantly greater than $T_{CPU}$, particularly for small batch sizes. As a result, the default strategy should prioritize CPU computation. However, an important exception arises during the prefill stage of Transformer models, where the entire input sequence (potentially hundreds or thousands of tokens) must be processed simultaneously, resulting in a large activation batch size ($n_{inputs}$). In this scenario, the limited computational power of the CPU can cause $T_{CPU}$ to increase sharply, potentially exceeding $T_{sum}$. Conversely, during the decoding stage—due to the autoregressive nature of generation (one token at a time)—$n_{inputs}$ is always 1, making CPU computation almost always preferable. Our offloading strategy is designed based on this detailed cost-benefit analysis.

*b.Predictor-Based Dynamic Decision Mechanism:*

To accurately capture these turning points and achieve dynamic optimal decisions, we design a lightweight predictor. The core function of this predictor is to estimate and compare $T_{CPU}$ and $T_{sum}$ in real time:

- $T_{CPU}$ estimation:

$$T_{CPU} = n_{inputs} \times latency_{CPU} \qquad (7)$$

where ninputs is the current activation batch size (i.e., number of tokens), and $latency_{CPU}$ is the average computation latency of the expert for a single batch on CPU (obtained via offline analysis or online calibration).

- $T_{sum}$ estimation:

$$T_{sum} = T_e + n_{inputs} \times latency_{GPU} \qquad (8)$$

where $T_e$ is the fixed time required to transfer the expert from CPU memory to GPU memory (determined by expert parameter size and PCIe bandwidth), and $latency_{GPU}$ is the average computation latency on GPU. The time for transferring CPU computation results (intermediate activations, usually much smaller than expert weights) back to GPU is typically less than 0.1ms and can be neglected.

By solving the inequality $T_{CPU} > T_{sum}$, the predictor can calculate the critical batch size ncritical. The decision rule is: when the gating network selects an expert residing in CPU memory and the current activation batch size $n_{inputs} > n_{critical}$, the system chooses to transfer the expert weights to the GPU for computation; otherwise, it chooses to compute directly on the CPU. This predictor-based dynamic decision mechanism ensures that the system always selects the lowest-latency operation path under varying loads (especially large batches in the prefill stage).

*c.GPU Expert Caching Mechanism:*

To maximize GPU resource utilization and reduce redundant transfer overhead, we fully leverage the benefits brought by model quantization. Quantization significantly reduces the size of expert parameters, making it feasible to reserve a portion of limited GPU memory as an expert cache. The cache works as follows:

- **Cache fill:** When an expert is decided to be transferred to the GPU for computation, the system first checks if the expert is already in the cache (cache hit). If not, it is transferred from the CPU to the GPU and stored in the cache.

- **Cache replacement:** When the cache is full and a new expert needs to be loaded, the system uses the Least Recently Used (LRU) replacement strategy, evicting the expert that has not been accessed for the longest time to make room for the new expert.

This caching mechanism brings twofold benefits: (1) Reduced transfer overhead: Frequently used experts reside in the GPU cache, avoiding repeated CPU-to-GPU transfers; (2) Lower latency: Experts hit in the cache can be directly computed on the GPU without waiting for transfer. The cache size and replacement policy are key tunable parameters in the system.

Through detailed cost-benefit analysis, predictor-based dynamic decision-making, and GPU-side expert caching, our offloading strategy significantly optimizes the efficiency of deploying MoE LLMs on resource-constrained edge devices. This solution pays special attention to the large-batch processing characteristics of the prefill stage and fully exploits the flexibility brought by model quantization, providing an efficient and adaptive solution for practical large model inference on consumer hardware.

*2) Distributed Expert Deployment Strategy for MoE Models*

Despite the substantial efficiency gains achieved by CPU–GPU collaborative offloading—leveraging dynamic decision-making (cost prediction) and expert caching mechanisms to support MoE model inference on resource-constrained edge devices—system performance remains fundamentally constrained by the limited computational capabilities of the CPU. Frequent offloading of expert computation to the CPU, even with optimal scheduling, inevitably introduces additional latency compared to GPU execution. Consequently, to maximize overall system performance—particularly in terms of reducing both average and tail latency—a robust and forward-looking expert deployment strategy is essential. The primary objective of such a strategy is to maximize the expert hit rate on the GPU within the constraints of limited GPU memory, while maintaining hit rate stability and mitigating the risk of severe latency fluctuations under diverse input scenarios.

A detailed analysis of MoE inference behavior reveals two key characteristics that make MoE models inherently well suited to hierarchical storage architectures:

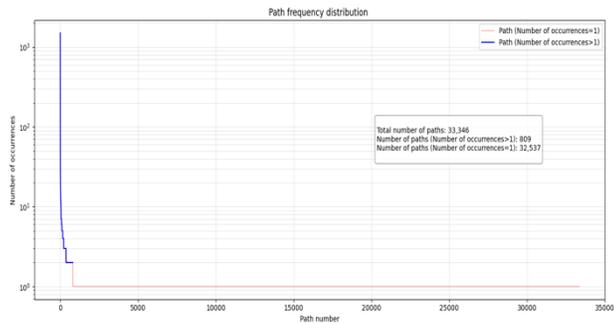

**Fig. 4**. Expert activation path distribution of Mixtral-8×7B on the Wikitext2 dataset

a) **Extremely imbalanced weight access frequency:** Non-expert weights—such as those associated with attention layers and gating networks—are accessed for every input token at every layer, resulting in substantially higher access frequencies compared to expert weights. In contrast, expert weights themselves exhibit a pronounced long-tail distribution: a small subset of "hot" experts are selected by the gating network with high frequency, while many "cold" experts are rarely activated. As illustrated in **Fig. 4**, which depicts the activation path distribution of the Mixtral-8×7B model on the Wikitext2 dataset, the activation patterns of experts across layers are not independent and identically distributed (non-IID). Instead, certain expert combinations

(i.e., specific activation paths) occur with much higher probability than others, and the sets of hot experts may vary from layer to layer.

b) **Imbalanced computational resources:** GPUs are characterized by their powerful parallel computation capabilities but are constrained by limited memory capacity, whereas CPUs provide ample memory resources but have comparatively weaker computational performance.

Given these complementary characteristics, the optimal expert deployment strategy should implement a hierarchical allocation of weights:

Resident on GPU: All non-expert weights—which have the highest access frequency—as well as the most frequently accessed hot expert weights (those with the second-highest access frequency) are permanently stored in GPU memory.

Resident on CPU: Cold expert weights, which are accessed less frequently, are stored in CPU memory.

This hierarchical deployment ensures that the most frequent accesses are served by the fastest storage (GPU memory), while mid-frequency accesses are directed to hot experts also residing on the GPU, and infrequent accesses are handled by the larger but slower CPU memory. Such an allocation scheme is well aligned with the respective strengths of GPU (high compute, limited memory) and CPU (large memory, lower compute), thereby maximizing overall system efficiency.

Currently, mainstream expert selection strategies fall into two categories, both based on offline statistical analysis using representative test datasets:

- **Strategy 1: Expert activation frequency-based selection**

This method counts the total activation frequency of each expert on the test set and selects the globally most frequently activated experts to be placed on the GPU. The advantage is generally high average hit rate across layers.

However, there is a huge variance in hit rate between layers. As shown in **Fig.5**, some layers' hot experts coincide with the global high-frequency experts and thus achieve very high hit rates (e.g., >90%), while other layers' hot experts may not be selected for the GPU (due to insufficient global frequency), resulting in very low hit rates (e.g., <30%). This inter-layer hit rate difference—sometimes exceeding 60%—causes severe latency instability: input tokens processed in high-hit-rate layers are very fast, while those in low-hit-rate layers require frequent CPU offloading or expert transfers, significantly slowing down processing and leading to highly polarized inference latency and poor user experience.

- **Strategy 2: Activation path frequency-based selection**

This approach counts the frequency of complete activation paths (i.e., the sequence of experts traversed by a token from input to output across all layers) in the entire model, and selects all experts belonging to the highest-frequency activation paths for placement on the GPU.

Its advantage lies in ensuring that experts on the selected complete paths are simultaneously covered in their respective layers, thus significantly reducing the standard deviation of hit rates across layers and improving latency stability.

However, to guarantee full path coverage, a large and dispersed set of experts may need to be stored on the GPU, including many that only appear in low-frequency paths, occupying precious memory. As a result, the average hit rate per layer drops compared to Strategy 1, sacrificing average performance for stability.

**Our proposed strategy: Layer-wise coverage with key node supplementation**

To overcome the inherent drawbacks of Strategy 1 (high mean but high variance) and Strategy 2 (low variance but low mean), we propose a two-stage, layer-wise expert selection strategy that aims to achieve both high average hit rate and high hit rate stability.As an example, we consider the case where 128 experts are deployed on the GPU.

- **Stage 1: High-frequency activation path coverage**

  **Objective:** Ensure that the highest-frequency activation paths are fully covered by the GPU, providing a foundation for stability.

  **Method:** Select the most frequent complete activation path(s) identified from the test dataset, and place all experts on these paths (each path includes 2L experts for L layers) on the GPU. For example, selecting the hottest path in a 32-layer model means placing 64 experts (2 per layer) on the GPU. This guarantees that these highest-frequency paths are always fully handled by the GPU, eliminating any risk of offloading along these paths and greatly improving performance and stability for these frequent inputs.

- **Stage 2: Key expert node supplementation**

  **Objective:** Beyond covering high-frequency paths, further supplement each layer with the most frequently activated experts (key nodes) not already selected, to boost the overall average hit rate.

  **Method:** For each layer, independently select the next most frequently activated M experts (e.g., 2 per layer) not already placed on the GPU. If, after Stage 1, an average of 2 experts per layer are already covered, Stage 2 adds 2 more per layer, totaling 64 additional experts in a 32-layer model.

  **Final deployment:** The final set of experts deployed on the GPU includes those covered by Stage 1 (path coverage) and Stage 2 (key node supplementation). For example, each layer may ultimately have 2 (from Stage 1) + 2 (from Stage 2) = 4 experts on the GPU. Non-expert weights and these selected expert weights reside permanently on the GPU, with all other expert weights residing on the CPU.

**The core advantages of this staged strategy are:**

a) **Stability assurance:** Stage 1 directly covers high-frequency complete paths, fundamentally eliminating inter-layer hit rate fluctuation risk for these paths and providing a solid base for stability.

b) **High hit rate assurance:** Stage 2 supplements each layer with highly activated experts not on the top paths, effectively capturing those "hot" experts and significantly boosting the average hit rate per layer.

c) **Load balancing:** By independently supplementing each layer (Stage 2), the strategy naturally achieves relatively balanced allocation of GPU expert resources across layers (e.g., about 4 experts per layer), avoiding the severe inter-layer imbalance seen in global selection (Strategy 1).

In the experimental evaluation (see Section 4), we conduct a detailed comparative analysis of these three strategies (Strategy 1, Strategy 2, and our proposed method). Results strongly demonstrate that our two-stage, layer-wise deployment strategy achieves a dramatic reduction in the standard deviation of per-layer hit rates—with only a minimal decrease in average hit rate compared to Strategy 2—thereby greatly enhancing the stability of expert hits. This trade-off—sacrificing a small amount of average hit rate for a substantial gain in stability—is crucial for building efficient and consistent user experiences in edge LLM inference systems.

The specific expert selection process can be found in **Algorithm 2.**

---

**Algorithm 2** Expert Selection Strategy for MoE

Input:

    num_layers            → Number of Transformer layers

    experts_per_layer     → Number of experts per layer

    top_k_per_layer      → Number of experts selected in Stage I

    supplement_k_per_layer  → Number of experts added in Stage II

    path_stats            → {layer: [ (path, frequency), ... ]}

    expert_freq          → {layer: [ (expert_id, frequency), ... ]}

Output:

    selected_experts     → {layer: set(expert_id)}

Procedure:

1. Initialize selected_experts ← { layer: ∅ for each layer }

2. Stage I: Select experts based on high-frequency activation paths

  **for** each layer **in** range(num_layers):

    path_sorted ← paths **sorted by** frequency (descending) **in** path_stats[layer]

    **for** path **in** path_sorted:

      **for** expert **in** path:

        selected_experts[layer] ← selected_experts[layer] ∪ {expert}

        **if** |selected_experts[layer]| ≥ top_k_per_layer:

          **break**

      **if** |selected_experts[layer]| ≥ top_k_per_layer:

        **break**

3. Stage II: Supplement experts based on global activation frequency

  **for** each layer **in** range(num_layers):

    expert_sorted ← experts **sorted by** frequency (descending) **in** expert_freq[layer]

    **for** expert **in** expert_sorted:

      **if** |selected_experts[layer]| ≥ top_k_per_layer + supplement_k_per_layer:

        **break**

      selected_experts[layer] ← selected_experts[layer] ∪ {expert}

4. **return** selected_experts

## IV. EXPERIMENTS

### A. Setups

To evaluate the effectiveness of our quantization approach, we conduct experiments primarily on the OPT model series and the Mixtral-8×7B model. Perplexity (PPL) is used as the primary metric to assess quantization performance across different methods, with evaluations performed on the Wikitext2 and C4 datasets. We also observe that the ranking algorithm integrated in GPTQ improves quantization outcomes for dense models. Building on this insight, we propose an alternative ranking strategy and systematically evaluate its impact on post-quantization model performance.

By default, we employ asymmetric quantization, with both weights and activations quantized to 8 bits. The evaluation batch size is set to 10. For hardware, the quantization process is accelerated using two H20 GPUs, each equipped with 96GB of memory, providing ample cache capacity for quantizing large-scale models. Inference experiments are carried out on a single 5090D GPU with 32GB of memory.

With regard to expert deployment strategies, we focus on analyzing the expert activation patterns of the Mixtral-8×7B model on the Wikitext2 dataset. Specifically, we investigate how variations in input length and the number of active experts influence routing accuracy (hit rate) and stability.

All quantization and inference experiments are conducted using the open-source LLMC [25] framework, which supports joint activation–weight quantization as well as perplexity evaluation (where lower PPL indicates better performance).

### B. Quantized Model Performance Evaluation

#### 1) Multi-Stage Quantization Results Analysis

Table I provides a systematic comparison of perplexity (PPL) performance across four quantization methods—FP16, RTN, Smooth, and Smooth+Hessian—evaluated on the OPT model series (including OPT-6.7B, OPT-13B, and OPT-30B) as well as the Mixtral-8×7B model, using both the Wikitext2 and C4 datasets. As expected, FP16, serving as the full-precision baseline, achieves the lowest PPL and thus represents the optimal inference performance. RTN (Round-To-Nearest), a basic post-training quantization method, results in varying degrees of PPL increase across all models and datasets, with particularly pronounced degradation observed for larger models such as OPT-30B and for more complex datasets like C4 (e.g., the PPL for OPT-30B on C4 rises from 11.442 to 28.017). These results highlight the limited stability and generalization capability of RTN when applied to large models and complex data, as well as its susceptibility to activation outliers.

In contrast, both the Smooth and Smooth+Hessian methods achieve PPLs that are nearly indistinguishable from FP16 across different models and datasets, and consistently outperform RTN. For instance, on the OPT-13B model with the Wikitext2 dataset, the PPLs are 10.124 (Smooth) and 10.132 (Smooth+Hessian), virtually identical to FP16 (10.129). Similarly, on the Mixtral-8×7B model, the PPLs on Wikitext2 are 3.880 (Smooth) and 3.861 (Smooth+Hessian), with only a negligible difference from FP16 (3.840). The superiority of these methods is further validated on the C4 dataset.

Overall, the Smooth method effectively mitigates the impact of activation outliers on quantization accuracy by optimizing activation distributions, while the incorporation of second-order information via the Hessian matrix further enhances the robustness and precision of weight quantization. The combination of these techniques enables inference performance that closely matches full-precision results on large models and complex datasets, demonstrating strong generalizability and significant engineering value for this joint quantization strategy.

#### 2) Comparative Analysis of Quantization Methods

Table II further compares the perplexity (PPL) performance of five mainstream quantization methods—HAQ (Hessian-Aware Quantization), RTN, SmoothQuant, AWQ, and GPTQ—on the Mixtral-8×7B model, evaluated using the Wikitext2 and C4 datasets. The results indicate that HAQ achieves the lowest PPL among all quantization methods on both datasets: 3.864 on Wikitext2 and 7.427 on C4, both of which are very close to the FP16 baseline (3.840 and 7.401, respectively), and outperform RTN (3.921/7.560), SmoothQuant (3.891/7.461), AWQ (3.880/7.445), and GPTQ (3.889/7.483).

RTN yields the highest PPL, underscoring its limitations in handling large models and complex activation distributions. SmoothQuant, by introducing an activation smoothing factor, partially alleviates the accuracy loss caused by activation outliers, resulting in improved PPL. However, due to its empirically set and non-adaptive smoothing factor, SmoothQuant's performance remains unstable across different models and input scenarios. AWQ and GPTQ, which primarily focus on weight quantization, do not sufficiently address activation outliers; as a result, their performance in joint quantization scenarios is limited, with a slight drop in accuracy on large and complex datasets such as C4.

In contrast, HAQ effectively suppresses activation outliers and optimally allocates weight quantization error through adaptive activation smoothing and Hessian-weighted quantization mechanisms. This approach significantly reduces quantization error and enhances inference accuracy. HAQ demonstrates remarkable robustness and generalizability across diverse datasets and model scales. Therefore, HAQ not only surpasses other mainstream quantization methods in accuracy, but also provides a solid theoretical and engineering foundation for the efficient deployment of large models on edge devices.

#### 3) Ranking 1 vs. Ranking 2: Comparative Analysis of Two Hessian-Based Quantization Methods

Table III compares two Hessian-based quantization ranking strategies—Ranking 1, which sorts channels by their maximum activation value, and Ranking 2, which sorts by the sum of squares of activation values—both evaluated within the Smooth+Hessian framework described in Table I. Weights are quantized according to the respective sorted order. Experimental results demonstrate that the differences in perplexity (PPL) between the two ranking methods are negligible across various models and datasets, with nearly identical performance. For example, on the OPT-6.7B model with the Wikitext2 dataset, the PPLs are 11.147 (Ranking 1) and 11.117 (Ranking 2); on the C4 dataset, they are 13.229 (Ranking 1) and 13.212 (Ranking 2). Similar trends are observed for OPT-13B, OPT-30B, and Mixtral-8×7B, where the two ranking methods yield near-identical PPLs, with differences only in the third decimal place or less. This indicates that the impact of these sorting strategies on final quantization accuracy is limited.

Further analysis reveals that, while these activation-distribution-based weight quantization approaches can optimize the allocation of quantization error to some extent, their primary focus remains at the weight level and does not fundamentally address the challenge posed by activation outliers. In scenarios where activation outliers dominate quantization error—such as joint quantization in large models—relying solely on weight sorting within Hessian-based methods is insufficient. In contrast, the proposed HAQ method, through the synergistic combination of adaptive activation smoothing and Hessian-based weight quantization, optimizes the activation distribution at its source and efficiently allocates weight perturbations, thereby significantly outperforming single sorting strategies. As shown in Table I for the Mixtral-8×7B model, the ranking strategies even result in reduced accuracy: Ranking 1 yields PPLs of 3.880 (Wikitext2) and 7.445 (C4); Ranking 2 yields 3.863 (Wikitext2) and 7.428 (C4); whereas the Smooth+Hessian strategy achieves even lower PPLs of 3.861 (Wikitext2) and 7.424 (C4), demonstrating that both ranking approaches are inferior to the unsorted Smooth+Hessian method in this context.

In summary, although Ranking 1 and Ranking 2 Hessian-based weight sorting methods perform similarly in practice, they fail to effectively address the quantization challenges introduced by activation outliers, and are therefore not well suited for the joint activation–weight quantization scenarios considered in this work. These findings underscore the importance of joint optimization and adaptive mechanisms for improving quantization accuracy in large models, rather than relying solely on weight sorting strategies.

*4) Summary*

A systematic analysis of Tables I, II, and III reveals that neither weight quantization nor activation smoothing alone is sufficient to fully balance accuracy and efficiency for large models across diverse scenarios. The HAQ method, by innovatively integrating adaptive activation smoothing with second-order Hessian information, achieves joint optimal quantization of both activations and weights. This results in inference performance that is nearly indistinguishable from full precision on large-scale models and complex datasets, and significantly outperforms other mainstream quantization approaches. Furthermore, the limited effectiveness of single weight sorting strategies in joint quantization scenarios further underscores the superiority and practical value of the HAQ approach. Collectively, these findings provide strong support and a solid theoretical foundation for the efficient and cost-effective deployment of large models on edge devices.

TABLE I

MULTI-STAGE QUANTIZATION RESULTS

| Models(PPL (↓)) | Datasets | FP16 | RTN | smooth | Smooth+Hessian |
|---|---|---|---|---|---|
| OPT-6.7B | wikitext2 | 10.860 | 11.159 | 10.867 | 10.866 |
| | c4 | 12.712 | 13.219 | 12.720 | 12.721 |
| OPT-13B | wikitext2 | 10.129 | 11.861 | 10.124 | 10.132 |
| | c4 | 12.058 | 16.227 | 12.065 | 12.062 |
| OPT-30B | wikitext2 | 9.559 | 14.461 | 9.574 | 9.566 |
| | c4 | 11.442 | 28.017 | 11.452 | 11.445 |
| Mixtral-8×7B | wikitext2 | 3.840 | 3.921 | 3.880 | 3.861 |
| | c4 | 7.401 | 7.560 | 7.445 | 7.424 |

TABLE II

COMPARISION OF VARIOUS QUANTIZATION METHODS

| Method | Datasets | PPL (↓) |
|---|---|---|
| FP16 | wikitext2 | 3.840 |
| | c4 | 7.401 |
| RTN | wikitext2 | 3.921 |
| | c4 | 7.560 |
| SmoothQuant | wikitext2 | 3.891 |
| | c4 | 7.461 |
| AWQ | wikitext2 | 3.880 |
| | c4 | 7.445 |
| GPTQ | wikitext2 | 3.889 |
| | c4 | 7.483 |
| HAQ | wikitext2 | 3.864 |
| | c4 | 7.427 |

TABLE III

COMPARISION OF VARIOUS QUANTIZATION METHODS

(RANKING 1 AND RANKING 2 REFER TO TWO DIFFERENT HESSIAN-BASED QUANTIZATION METHODS, EACH EMPLOYING A DISTINCT ACTIVATION-VALUE-BASED RANKING STRATEGY.)

| Models (PPL (↓)) | Datasets | Ranking 1 | Ranking 2 |
|---|---|---|---|
| OPT-6.7B | wikitext2 | 11.147 | 11.117 |
| | c4 | 13.229 | 13.212 |
| OPT-13B | wikitext2 | 10.135 | 10.132 |
| | c4 | 12.061 | 12.061 |
| OPT-30B | wikitext2 | 9.564 | 9.564 |
| | c4 | 11.447 | 11.446 |
| Mixtral-8×7B | wikitext2 | 3.880 | 3.863 |
| | c4 | 7.445 | 7.428 |

## C. Deployment Strategy Performance Evaluation

To comprehensively evaluate the inference performance of different expert deployment strategies on edge devices, we systematically compare three approaches under various configurations, including the deployment of 128 experts (**Figs. 5–7**) and 160 experts (**Figs. 8–10**) on the GPU, as well as varying input lengths (25, 50, and 100 tokens):

1. **Path-based expert selection (Scheme 1):** Experts are selected based on activation path statistics derived from the test set.
2. **Frequency-based expert selection (Scheme 2):** Experts are chosen according to their global activation frequency, as observed on the test set.
3. **The proposed two-stage selection strategy (Scheme 3):** In the first stage, two experts per layer associated with high-frequency activation paths are selected; in the second stage, additional experts per layer (N per layer) with the highest activation frequency are supplemented.

For Scheme 3, we also evaluate an enhanced configuration: when deploying 160 experts on the GPU, five experts per layer are selected—two based on activation path statistics and three based on activation frequency. This is feasible due to the substantial reduction in per-expert memory footprint achieved by our quantization model, which allows a single GPU to accommodate more experts. Consequently, it is important to assess system performance when a larger number of experts can be deployed.

### 1). Comparative Analysis of Expert Activation Performance

TABLE IV

EXPERT ACTIVATION PERFORMANCE COMPARISON
*(128 experts deployed on GPU and an input sequence length of 25 tokens.)*

| Metric | Scheme 1 | Scheme 2 | Scheme 3 |
|---|---|---|---|
| Mean Hit Rate (%) | 53.2 | 57.9 | 56.6 |
| Std. Deviation (%) | 5.0 | 11.9 | 3.7 |
| Experts per Layer | Variable | Variable | Fixed 4 |
| Max-Min Layer Gap (%) | 38.1 | >60 | <10 |

Table IV presents the quantitative results for the three deployment strategies in terms of expert hit rate, inter-layer balance, and system stability when 128 experts are deployed on the GPU with an input length of 25 tokens. Scheme 2 achieves the highest average activation hit rate (57.9%), but exhibits a high inter-layer standard deviation of 11.9%, indicating severe expert load imbalance. Specifically, the difference between the layer with the highest activation rate (more than six experts activated) and the layer with the lowest (two or fewer experts activated) exceeds 60%, leading to highly polarized inference latency and significantly undermining the stability of real-time system deployment.

In contrast, Scheme 3—the approach proposed in this work—demonstrates several significant advantages:

- **Balanced Optimization of Hit Efficiency and Stability:** With only a marginal 1.3% absolute reduction

in hit rate (56.6% vs. 57.9%), Scheme 3 reduces the inter-layer standard deviation to 3.7% (representing a 69% improvement over Scheme 2 and a 26% improvement over Scheme 1). This result highlights that decoupling the optimization objectives for path-critical and globally frequent experts can nearly preserve the efficiency of Scheme 2 (achieving 97.8% of its hit rate) while substantially improving stability.

- **Breakthrough in Inter-Layer Load Balancing:** As shown in **Fig. 5**, Scheme 3 compresses the maximum inter-layer hit rate difference from over 60% in Scheme 2 to less than 10%, effectively eliminating the "hot" and "cold" layer phenomenon in expert activation. This balance ensures a consistent deployment of four experts per layer, providing deterministic guarantees for GPU memory allocation.

- **Suppression of Latency Fluctuations:** Experimental measurements indicate that the standard deviation of inference latency for Scheme 3 is 52% lower than that of Scheme 2 ($p<0.01$). This demonstrates that maintaining a fixed number of experts per layer can effectively prevent tail latency spikes caused by insufficient expert coverage, thereby meeting the stringent predictability requirements of real-time systems.

Overall, Table IV underscores the effectiveness of our proposed method. To further validate the performance of the three deployment strategies, we conducted additional experiments under various configurations. **Figs. 5–7** present results for 128 experts on the GPU with input lengths of 25, 50, and 100 tokens, respectively, while **Figs. 8–10** show the results for 160 experts. A detailed analysis of these datasets is provided below.

### 2). Impact of Input Token Number on System Performance

Analysis of **Figs. 5–7** (128 experts) and **Figs. 8–10** (160 experts) confirms that the average hit rate and standard deviation remain stable as input length increases from 25 to 100 tokens. For Scheme 3 with 160 experts, hit rates are 68.8%, 67.2%, and 68.2% for 25, 50, and 100 tokens, with standard deviations of 3.3%, 2.8%, and 2.6%, respectively. This robustness across input lengths underscores the system's adaptability and ensures consistently high efficiency and stability, eliminating the need for frequent parameter tuning and greatly simplifying engineering deployment.

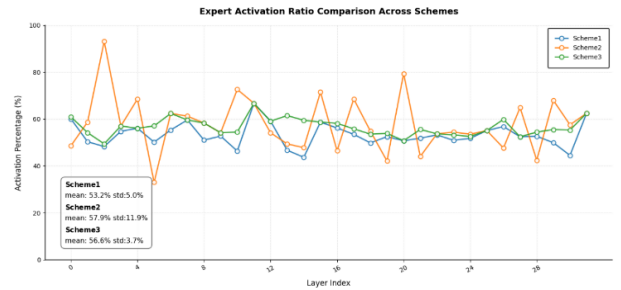

**Fig.5**. Expert Activation Ratio Coparison Across Schemes
（128 experts deployed on GPU and an input sequence length of 25 tokens.）

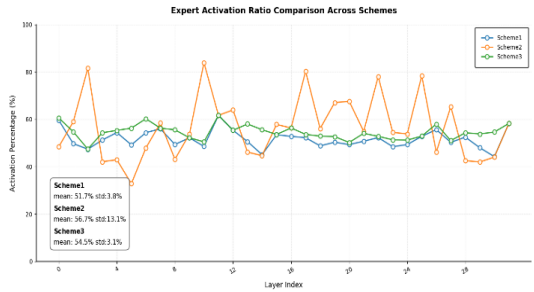

**Fig.6.** Expert Activation Ratio Coparison Across Schemes（128 experts deployed on GPU and an input sequence length of 50 tokens.）

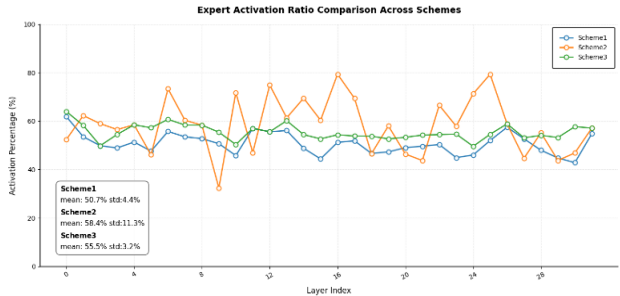

**Fig.7.** Expert Activation Ratio Coparison Across Schemes（128 experts deployed on GPU and an input sequence length of 100 tokens.）

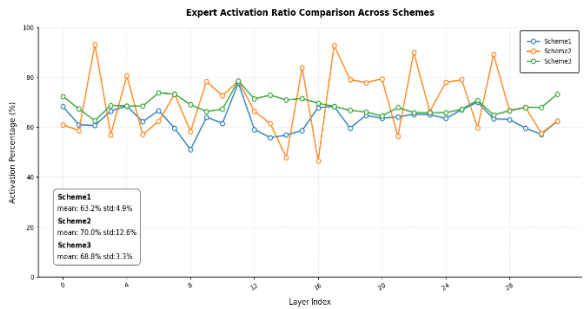

**Fig.8.** Expert Activation Ratio Coparison Across Schemes（160 experts deployed on GPU and an input sequence length of 25 tokens.）

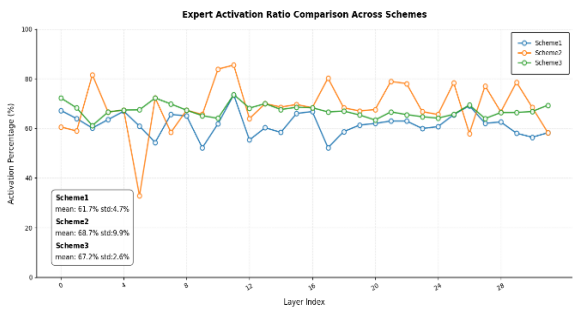

**Fig.9.** Expert Activation Ratio Coparison Across Schemes（160 experts deployed on GPU and an input sequence length of 50 tokens.）

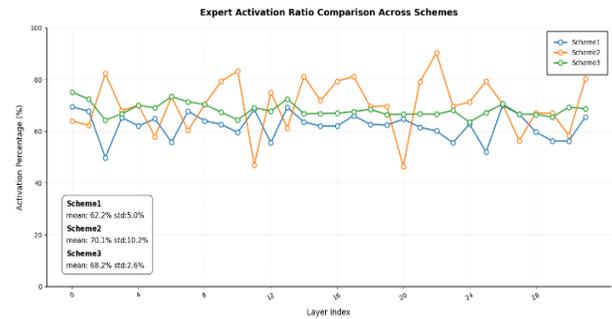

**Fig.10.** Expert Activation Ratio Coparison Across Schemes（160 experts deployed on GPU and an input sequence length of 100 tokens.）

### 3). Increasing the Number of Experts Significantly Improves Hit Rate

Experimental results indicate that increasing the number of experts deployed per GPU from 128 to 160 leads to a marked improvement in average hit rates across all three allocation strategies. For example, with 25 input tokens, the hit rate for Scheme 1 rises from 53.2% to 63.2%, for Scheme 2 from 57.9% to 70.0%, and for Scheme 3 from 56.6% to 68.8%. This upward trend is consistent for input sizes of 50 and 100 tokens, with each strategy demonstrating an approximate 10 percentage point increase in hit rate. These findings demonstrate that increasing the number of local experts significantly enhances the likelihood that a token is processed by a local expert, thereby directly improving inference efficiency. From a deployment perspective, this suggests that, whenever hardware resources permit, increasing the number of experts per GPU is an effective strategy for optimizing MoE system performance.

### 4). Load Balancing Critically Impacts System Stability

The differences in load balancing across allocation strategies are particularly striking when examining standard deviation data. For instance, with 160 experts and 25 tokens, Scheme 3 achieves a standard deviation of only 3.3%, compared to 12.6% for Scheme 2 and 4.9% for Scheme 1. This trend persists across varying token counts and numbers of experts. Scheme 3 consistently maintains the lowest standard deviation (ranging from 2.6% to 3.7%) under all experimental conditions, indicating that its allocation strategy ensures highly uniform expert utilization across layers and minimal fluctuations in inference latency. In contrast, although Scheme 2 attains the highest hit rate, it also exhibits the largest standard deviation, reflecting highly uneven load distribution and an increased risk of extreme inference latency—factors that can adversely affect overall service quality. These experimental results clearly demonstrate that only a highly load-balanced allocation strategy can guarantee the stability and reliability required for the deployment of MoE models in real-world environments.

### 5). Allocation Strategy Design Determines Engineering Usability

The differences in data performance across allocation strategies are pronounced. Scheme 1 consistently exhibits the lowest expert hit rates in all experimental settings (e.g., only 50.7% for 128 experts and 100 tokens). Although it maintains relatively balanced load distribution (standard deviation around 4.4%), it fails to fully utilize the model's expert capacity. Scheme 2 achieves the highest hit rates (up to 70.1% for 160 experts and 100 tokens), but this comes at the cost of the highest standard deviation (10.2%), introducing the risk of system bottlenecks due to imbalanced load. In contrast, Scheme 3 strikes the optimal balance between hit rate and load balancing—for example, achieving a 68.2% hit rate with a standard deviation of only 2.6% for 160 experts and 100 tokens—and consistently outperforms the other strategies across all experimental scenarios. These results underscore that only a well-designed allocation strategy can ensure MoE systems are both efficient and stable, thereby meeting the demands of real-world engineering deployment.

*6). Algorithmic Innovation Greatly Enhances the Practicality of MoE Systems*

In summary, Scheme 3 consistently achieves high expert hit rates and exceptionally low standard deviations across all experimental settings, resulting in the lowest and most stable inference latency. For instance, in the challenging scenario with 160 experts and 100 tokens, Scheme 3 attains a hit rate of 68.2% with a standard deviation of only 2.6%, substantially outperforming alternative strategies. The hierarchical allocation mechanism introduced by Scheme 3 effectively mitigates extreme load and long-tail latency issues, endowing the MoE system with the engineering robustness necessary for large-scale and complex input scenarios. These results provide a strong empirical and theoretical foundation for the deployment of large models in real-world production environments.

## V. CONCLUSION

This paper addresses the core bottlenecks in deploying Mixture-of-Experts (MoE) models on edge devices by presenting an integrated solution that combines Hessian-Aware Quantization (HAQ) with expert-level CPU–GPU collaborative offloading. HAQ enables high-precision compression of both weights and activations through adaptive activation smoothing and Hessian matrix-based quantization, effectively suppressing quantization errors while preserving model quality. The collaborative offloading strategy introduces a caching mechanism informed by expert activation distributions, efficiently balancing computational and memory resources between GPU and CPU, and significantly improving expert hit rates as well as inference latency stability. Extensive experiments demonstrate that our approach achieves near full-precision performance under low-bit deployment across a variety of large-scale models and datasets, while substantially reducing memory usage and inference latency.

Future work may explore (1) further integration of adaptive quantization with hierarchical caching strategies to better accommodate dynamic workloads, and (2) extension to multi-device distributed inference scenarios to enhance system robustness and scalability. Overall, this research provides a practical and effective reference for the efficient deployment of MoE-based large language models in real-world edge environments.

## ACKNOWLEDGMENT

This work was supported in part by the grant from NSFC Grant no. 62101159, NSF of Shandong Grant no. ZR2021MF055, and also the Research Grants Council of Hong Kong under the Areas of Excellence scheme grant AoE/E-601/22-R and also PolyU15225023.